\documentclass[letterpaper, 10 pt, conference]{ieeeconf}  

\IEEEoverridecommandlockouts                              

\usepackage{amsmath} 
  \usepackage{amssymb, amsthm}  

\usepackage{todonotes}
\usepackage{siunitx}
\usepackage[hidelinks]{hyperref}
\usepackage[capitalize,nameinlink]{cleveref}
\usepackage{algorithm}
\usepackage{algpseudocode}
\usepackage{placeins}
\usepackage{float}

\usepackage{pgfplots}

\newtheorem{assumption}{\bf Assumption}
\newtheorem{theorem}{\bf Theorem}

\newtheorem{solution}{\bf Solution} 
\crefname{assumption}{Assumption}{Assumptions}
\crefname{theorem}{Theorem}{Theorems}
\crefname{solution}{Solution}{Solution}

\makeatletter
\let\MYcaption\@makecaption
\makeatother
\usepackage[font=footnotesize]{subcaption}
\makeatletter
\let\@makecaption\MYcaption
\makeatother

\pgfplotsset{compat=1.18}
\pgfplotsset{every axis/.append style={
    label style={font=\small},
    tick label style={font=\small}  
}}

\title{\LARGE \bf
AVOID-JACK: Avoidance of Jackknifing for Swarms of Long Heavy Articulated Vehicles
}

\author{Adrian Schönnagel$^{1,2}$, Michael Dub\'{e}$^{1}$, Christoph Steup$^{2}$, Felix Keppler$^{2}$ and Sanaz Mostaghim$^{1,2}$
\thanks{This research was funded by EFRE Saxony-Anhalt, grant number ZS/2023/12/182177, and supported by the Fraunhofer Internal Programs under Grant No. Attract 40-11882.}
\thanks{$^{1}$ Chair for Computational Intelligence, Otto-von-Guericke-University, Magdeburg, Germany {\tt\small \{adrian.schoennagel, michael.dube, sanaz.mostaghim\}@ovgu.de}}%
\thanks{$^{2}$ Fraunhofer Institute for Transportation and Infrastructure Systems IVI, Dresden, Germany
        {\tt\small \{christoph.steup, felix.keppler\}@ivi.fraunhofer.de}}%
}

\begin{document}

\maketitle
\thispagestyle{empty}
\pagestyle{empty}

\begin{abstract}

This paper presents a novel approach to avoiding jackknifing and mutual collisions in Heavy Articulated Vehicles (HAVs) by leveraging decentralized swarm intelligence. In contrast to typical swarm robotics research, our robots are elongated and exhibit complex kinematics, introducing unique challenges. Despite its relevance to real-world applications such as logistics automation, remote mining, airport baggage transport, and agricultural operations, this problem has not been addressed in the existing literature.

To tackle this new class of swarm robotics problems, we propose a purely reaction-based, decentralized swarm intelligence strategy tailored to automate elongated, articulated vehicles. The method presented in this paper prioritizes jackknifing avoidance and establishes a foundation for mutual collision avoidance. We validate our approach through extensive simulation experiments and provide a comprehensive analysis of its performance. For the experiments with a single HAV, we observe that for \SI{99.8}{\percent} jackknifing was successfully avoided and that \SI{86.7}{\percent} and \SI{83.4}{\percent} reach their first and second goals, respectively. With two HAVs interacting, we observe \SI{98.9}{\percent}, \SI{79.4}{\percent}, and \SI{65.1}{\percent}, respectively, while \SI{99.7}{\percent} of the HAVs do not experience mutual collisions.
\end{abstract}

\section{INTRODUCTION}
Distilling complex global swarm behaviors into simple, local rules for individuals is a fascinating challenge \cite{hamannSwarmRoboticsFormal2018}. This challenge is especially pronounced in swarm robotics, where robots, unlike biological organisms, are engineered to fulfill human-defined objectives. Swarm robotics has proven effective at a diverse range of tasks, enabling robot swarms to achieve collective goals through interactions governed by these local rules \cite{hamannSwarmRoboticsFormal2018}. 

In most models, robots are approximated as point masses subject to basic kinematic equations (cf. \cite{hamannSwarmRoboticsFormal2018,diasSwarmRoboticsPerspective2021}). Common examples include spherical agents, differential-drive vehicles, and aerial drones. Methods for coordinating their motion, whether forming swarm patterns, navigating to target locations, or collaborating on tasks, are widely documented \cite{diasSwarmRoboticsPerspective2021}.

In \cite{schonnagelSwarmsLongHeavy2025}, we introduced a new challenge: swarms of long, Heavy Articulated Vehicles (HAVs). Two novel subproblems emerge in this context: avoiding infeasible articulation states (i.e., \textit{jackknifing}) and space-efficient mutual collision avoidance. Solving these problems could unlock applications in decentralized conflict resolution for logistics and remote mining operations, as well as agricultural tasks such as crop harvesting or combined soil preparation and seeding, and scenarios that could use Automated Ground Vehicles (AGVs) with trailers, such as airport baggage transport and factory automation. In this paper, we focus on avoiding jackknifing in HAVs by introducing a novel attraction-repulsion-based swarm behavior referred to as \textit{AVOID-JACK}. \cref{sec:p_desc} details the problem formulation.


\section{State of the Art}

The literature on multi-robot systems is divided into centralized and decentralized approaches. Centralized methods for robots with simple kinematics are well established and extensively studied \cite{capPrioritizedPlanningAlgorithms2015a,sternMultiAgentPathfindingDefinitions2019}. However, research addressing robots with complex kinematics, such as HAVs, is much more limited. Some works have explored centralized coordination of trajectories for these platforms \cite{kepplerPrioritizedMultiRobotVelocity2020}, and a few have considered dynamic rescheduling in response to disturbances \cite{schaferRMTRUCKDeadlockFreeExecution2023}. Local path planning for such vehicles has been studied \cite{eluemeReviewMotionPlanningMethods2024} but is not yet fully integrated into multi-robot systems.

On the decentralized side, the field of swarm robotics has focused on robots with simple kinematic models \cite{hamannSwarmRoboticsFormal2018,diasSwarmRoboticsPerspective2021,engelbrechtFundamentalsComputationalSwarm2006}. There have been some extensions to non-holonomic systems, such as cars \cite{kumarLyapunovBasedControlSwarm2015} and fixed-wing aircraft \cite{hauertReynoldsFlockingReality2011}, among others. While these studies incorporate movement constraints, they often rely on simplified collision avoidance strategies and do not fully capture more complex kinematic behaviors.
\begin{figure}[b!]
    \centering
    \includegraphics{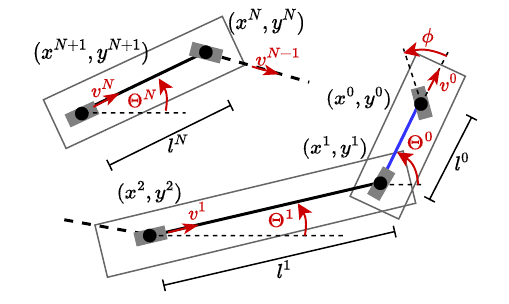} 
    \caption{Ackermann Truck-Trailer Model for HAV $i$. The truck (blue) and first trailer (black) are shown at the bottom right, while the final trailer $N_i$ is depicted at the top left. Intermediate trailers are omitted for clarity and indicated by a thick dashed line. For simplicity, the subscript $i$ is omitted from the variables.}
    \label{fig:kinematics}
\end{figure}

Some work exists in the area of distributed model predictive control~(MPC), but these either focus on trucks without trailers for arbitrary tasks~\cite{dmpc_truck} or handle trailers, but only for very specific tasks such as lane changing~\cite{dmpc_truck_trailer_platooning}. Even less work exists in the area of distributed multi-robot systems~(MRS) control, focusing on very specific use cases such as leader-follower behavior~\cite{mrs_distributed_control}.

To the best of our knowledge, fully decentralized approaches for multi-robot systems with complex kinematics remain an open research area and are the focus of this study.

%

\section{PROBLEM DESCRIPTION}
\label{sec:p_desc}

In decentralized swarms of long HAVs — each comprising an Ackermann-steered truck towing one or more passive trailers 
— the two major challenges are avoiding jackknifing and efficient mutual collision avoidance \cite{schonnagelSwarmsLongHeavy2025}.
These challenges persist regardless of the swarm’s specific objectives. 

\cref{sec:kinematics} details our vehicle assumptions and kinematic model. Sections \ref{sec:jackknifing} and \ref{sec:collision} examine the jackknifing and collision avoidance problems, respectively. Finally, \cref{sec:scenario} describes the scenario studied in this paper. 

\subsection{KINEMATIC MODEL}
\label{sec:kinematics}

We model each HAV $i$ as a truck pulling $N_i$ trailers, assuming negligible width, on-axle hitching, and length equal to wheelbase for each segment. Let a \textit{virtual-axle} denote the centroid of all corresponding physical axles. Then, let the truck’s front and rear virtual‐axles be indexed by $k=0,1$, respectively, and each trailer $j\in[1,N_i] \subset \mathbb{N}$ have a single rear virtual‐axle indexed by $k=j+1$. Let $(x_i^k,y_i^k)$ denote the position of the $k^{th}$ virtual-axle and let $\Theta_i^0,\Theta_i^j$ be the heading of the truck and trailer $j$, respectively.
The truck’s and $j^\text{th}$ trailer's wheelbases are $l_i^0$ and $l_i^j$, respectively.
A schematic of HAV $i$ with $N_i$ trailers is shown in \cref{fig:kinematics}.

Each HAV is controllable through the velocity $v_i^0\geq0$ and steering angle $\phi_i\in[-\phi_i^\text{max},\phi_i^\text{max}]$ of the truck. Where $v_i^0$ is the truck's speed, $\Theta_i^0$ is its direction, and $v_i^j$ is the $j^{th}$ trailer's speed.
With the nomenclature and on-axle hitching simplification, we adapt the differential equations from \cite{sordalenConversionKinematicsCar1993}, provided in \labelcref{eq:kinematics,eq:kinematics-trailers}.
    \begin{align}
        \begin{pmatrix}
            {\dot{x}_i^1}\\
            {\dot{y}_i^1}\\
            {\dot{\Theta}_i^0}
        \end{pmatrix} = \begin{pmatrix}
            {v_i^0 \cdot \cos(\Theta_i^0)}\\
            {v_i^0 \cdot \sin(\Theta_i^0)}\\
            {v_i^0/\mathrm{l_i^0} \cdot \tan(\phi_i)}
        \end{pmatrix}
        \label{eq:kinematics}
    \end{align}
    \begin{align}
        \forall j\in[1,N_i]:
        \begin{pmatrix}
            {\dot{\Theta}_i^j}\\
            {v_i^j}
        \end{pmatrix} = \begin{pmatrix}
            {-\frac{v_i^{j-1}}{\mathrm{l_i^j}} \cdot \sin(\Theta_i^{j}-\Theta_i^{j-1})}\\
            {v_i^{j-1} \cdot \cos(\Theta_i^{j}-\Theta_i^{j-1})}
        \end{pmatrix} 
        \label{eq:kinematics-trailers}
    \end{align}

\subsection{JACKKNIFING}
\label{sec:jackknifing}
Jackknifing occurs when the angle between two consecutive segments exceed a maximum articulation angle, leading to severe damage to the vehicle. We denote the articulation angle between trailer $j$ and its preceding segment as $\delta_i^j=\Theta_i^j-\Theta_i^{j-1}$ for $j=1,\dots,N_i$. Through trigonometric relationships, the inequality constraints \eqref{eq:delta_lim} become \eqref{eq:cos_delta_lim}. Although we set this angle limit to $\SI{90}{\degree}$ ($\pi/2$ radians), the equations work for any angle $\in(0,\pi]$.
\begin{align}
    \forall& j\in[1,N_i]:&  \pi/2 &\geq \Big|\delta_i^j\Big| \label{eq:delta_lim}\\
    \forall& j\in[1,N_i]:&  0 &\leq \cos{\bigl(\delta_i^j\bigr)} \label{eq:cos_delta_lim}
\end{align}

\subsection{MUTUAL COLLISION AVOIDANCE}
\label{sec:collision}
HAV $i$ can be approximated as a circle around the truck with center $(x_i^1,y_i^1)$ and \textit{collision radius} $d_i=\sum_{j=1}^{N_i}l_i^j$ (the \textit{footprint}). Then the \textit{collision distance} of HAV $i$ and $h$ is $d_{i,h}=d_i + d_h$. They are in \textit{potential collision} if their circles overlap, that is \labelcref{eq:potential-collision} holds. For a more exact version, consider $P_i$ to be a polygonal chain connecting the axles of HAV $i$. If the polygonal chain of HAVs $i$ and $h$ intersect, they are considered to be in \textit{actual collision} (i.e., \labelcref{eq:collision} holds).
\begin{align}
     \bigl\lVert (x_i^1,y_i^1)^T - (x_h^1,y_h^1)^T \bigr\rVert_2 &\le d_{i,h}
     \label{eq:potential-collision} \\
     P_i \cap P_h &\ne \emptyset
     \label{eq:collision}
\end{align}

\subsection{SCENARIO}
\label{sec:scenario}
Let there be $M$ HAVs in a 2D space with no known obstacles. Each HAV $i$ starts from a \textit{start pose} $\mathcal{P}_i^S=\{p_{x,i}^S,p_{y,i}^S,p_{\Theta,i}^S\}$ such that its truck's rear axle $(x_i^1,y_i^1)$ and heading $\Theta_i^0$ are assigned these values, and its trailers have no articulation. It is then assigned consecutive \textit{goal poses} $\mathcal{P}_i^G=\{p_{x,i}^G,p_{y,i}^G,p_{\Theta,i}^G\}$ for the same axle. In these, trailer articulation is not restricted except for jackknifing; see \labelcref{eq:cos_delta_lim}.

The start and goal poses for all HAVs are chosen such that they are not in potential collision; that is, there is no pair $(i,h) \in \{1,\dots,M\}^2, i\ne h$ for which \labelcref{eq:potential-collision} holds.

\section{AVOID-JACK}

\begin{figure}[ht]
    \centering
    \includegraphics{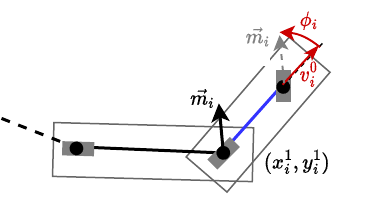} 
    \caption{Translation of Vector to Ackermann Movement.}
    \label{fig:movement}
\end{figure}

Let a vector $\vec{m_i} = \left(m_i^x,m_i^y\right)^T$ define the desired movement of HAV $i$. Then, one can translate $\vec{m_i}$ into the Ackerman control commands $v_i^0$ and $\phi_i$ by \labelcref{eq:movement-1,eq:movement-2}; see \cref{fig:movement}. Note that $\phi_i$ is clipped into the HAVs' steering limits $[-\phi_i^\text{max},\phi_i^\text{max}]$. 
\begin{align}
    v_i^0 &= \lVert\vec{m_i}\rVert_2 \label{eq:movement-1}\\
    \phi_i &= \arctan\bigl(m_i^x,m_i^y\bigr)-\Theta_i^0 \label{eq:movement-2} \\
            & \qquad \text{s.t. } \phi_i \in [-\phi_i^\text{max} ,\phi_i^\text{max}] \nonumber
\end{align} 
With this notation, attractive and repulsive forces on the vehicle are modeled as independent vectors and then combined into a total movement vector. Given the segment lengths $l_i^j$ and \labelcref{eq:kinematics,eq:kinematics-trailers}, the next vehicle state $\mathcal{S}_i=\{x_i^1,y_i^1,\Theta_i^0,\dots,\Theta_i^{N_i}\}$ follows. All axle positions $(x_i^k,y_i^k)$ can then be obtained via trigonometry if needed.

In the remainder of this section, we propose a method to model jackknife avoidance, goal attraction, and mutual collision avoidance via such attractive and repulsive vectors.


\subsection{JACKKNIFE AVOIDANCE}
\label{sec:jack}
Our jackknife avoidance strategy models alignment of the truck-trailers through repulsion forces, which increase with articulation. This is possible because the truck and its trailers are not individuals but are tied together, which translates tugging forces of the truck on the trailers to alignment changes. Our approach is presented in \cref{theo:jack,sol:jack} and then proven given \cref{ass:straight-keeps-straight,ass:straight-reduces-jackknifing}.
\begin{assumption}
    When an HAV is almost straight, a low or zero steering angle will make it straighter. \label{ass:straight-keeps-straight}
\end{assumption}
\begin{assumption}
    Aligning the truck with its first trailer and then continuing straight ahead will reduce articulation in all trailers in the long term. Although the articulation could increase in rearward joints for a short time horizon $\Delta_T$ it will decrease thereafter; cf. \labelcref{eq:ass2}.
    \label{ass:straight-reduces-jackknifing} 
    \begin{equation}
        \exists \Delta_T\geq0 : \forall t\geq0,j: \bigl|\delta_i^j(t_0+\Delta_T+t)\bigr| \leq \bigl|\delta_i^j(t_0)\bigr|
        \label{eq:ass2}
    \end{equation}
\end{assumption}
\begin{theorem}\label{theo:jack}
    A properly designed behavior consisting of a \\weight function $w^\text{jack}(\delta)$ combining repulsion $(\cos \Theta_i^1, \sin \Theta_i^1)^T$ and goal attraction $\vec{m}_i^G$ will repel the HAV from non-recoverable articulation states.
\end{theorem}
\begin{solution}[to \cref{theo:jack}]\label{sol:jack}
    The weight function $w^\text{jack}(\delta)$ is designed as the activation of the articulation angle repulsion; see \labelcref{eq:jackknife-weigt-func} and \cref{fig:jackknife-weight}. The weight is then calculated for each joint of the HAV, summed, and multiplied with the direction induced by the heading $\Theta_i^1$ of the first trailer, see \labelcref{eq:jackknife-vector-addition-tmp}. Given any goal heading vector $\vec{m}_i^g$ with length $\lVert\vec{m}_i^g\rVert_2=1$, one can obtain the total movement vector $\vec{m}_i$ through \labelcref{eq:jackknife-vector-addition-tmp} by \labelcref{eq:jackknife-vector-addition}.
\end{solution}
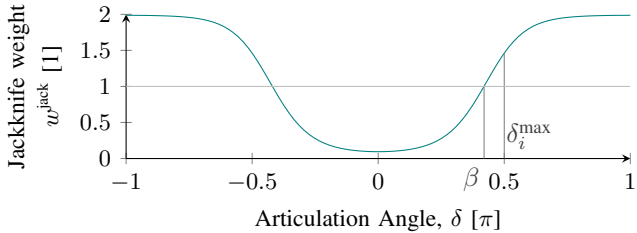
\begin{figure}[th]
    \centering
    \begin{tikzpicture}
        \begin{axis}[
            axis lines = left,
            xlabel={Articulation Angle, $\delta$ [$\pi$]},
            ylabel={\shortstack{Jackknife weight\\ $w^\text{jack}$ [1]}},
            xmin=-1, xmax=1,
            ymin=0, ymax=2,
            clip=false,
            height=3.5cm,
            width=\linewidth-10pt
        ]
        \addplot [
            domain=-1:1, 
            samples=100, 
            color=teal,
        ]
        {1 + tanh(1 * (0.5 - 2 * cos(deg(pi*\x))))};

        \def\Xone{0.42}
        \def\Xtwo{0.5}

        \addplot[gray] coordinates {(\Xone,0) (\Xone,1)};
        \addplot[gray] coordinates {(\Xtwo,0) (\Xtwo,1.46)};
        \node[below, darkgray] at (axis cs:\Xone-0.05,0) {$\beta$};
        \node[above, darkgray] at (axis cs:\Xtwo+0.1,0) {$\delta_i^\text{max}$};

        \addplot[lightgray] coordinates {(-1,1) (1,1)};
        
        \end{axis}
    \end{tikzpicture}
    \caption{Function for Jackknife Weight over Articulation Angle, see \labelcref{eq:jackknife-weigt-func}.}
    \label{fig:jackknife-weight}
\end{figure}
\begin{align}
    \vec{u}_i &= \vec{m}_i^g + 
    \left[\sum_{j=1}^{N_i} w^\text{jack}\left(\delta_i^j\right)\right] \label{eq:jackknife-vector-addition-tmp}
    \cdot\begin{pmatrix}
        \cos\Theta_i^1 \\ \sin\Theta_i^1
    \end{pmatrix} \\
    \vec{m}_i &= v_i^\text{max} \cdot \frac{\vec{u}_i}{\lVert\vec{u}_i\rVert_2} \label{eq:jackknife-vector-addition}
\end{align}

\begin{proof}[Proof of \cref{theo:jack} using \cref{sol:jack}] \label{sec:jack-proof}
    Starting from the straight start pose $\mathcal{P}_i^S$, it is obvious that with $\phi_i\rightarrow0$ all $|\delta_i^j|$ will decrease, and that with $|\phi_i|\rightarrow\phi_i^\text{max}$ articulation will increase the most. Then, the articulation of the first trailer will increase more than that of consecutive trailers. That is $\forall j: \dot\delta_i^1\geq\dot\delta_i^j$ since $v_i^0\geq v_i^j$; see \labelcref{eq:kinematics-trailers}. 
    
    Through the design of $w^\text{jack}(\delta)$ one can choose $\beta<\delta_i^\text{max}$. Through case distinction, we show stability when starting from a straight HAV:
    \begin{itemize}
        \item If \emph{$|\delta_i^1|>\beta$} then $w_i^\text{jack}>1$. Therefore, all $\delta_i^j$ will decrease in absolute value in the next step.
        \item If \emph{$|\delta_i^1|=\beta$} then $|\Delta\delta_i^1| < \delta_i^\text{max}-\beta$ must hold in $\Delta t$ in order to not jackknife. Combing $\dot\delta_i^1 = \dot\Theta_i^1-\dot\Theta_i^0$ with \labelcref{eq:kinematics,eq:kinematics-trailers} yields \labelcref{eq:delta_dot}:
        \begin{align}
            \dot\delta_i^1 = -\frac{v_i^0}{l_i^1}\sin(\delta_i^1)-\frac{v_i^0}{l_i^0}\tan(\phi_i)
            \label{eq:delta_dot}
        \end{align}
        Over one time step of $\Delta t$ the maximum change in $\delta_i^1$ can then be computed through \labelcref{eq:delta_dot_max_tmp} by \labelcref{eq:delta_dot_max}:
        \begin{align}
            \max{|\Delta\delta_i^1|} &= \max{|\dot\delta_i^1|}\cdot\Delta t \label{eq:delta_dot_max_tmp}\\
            &= v_i^0\Delta t \cdot\underbrace{\Bigl|\frac{\sin(\delta_i^1)}{l_i^1}+\frac{\tan(\phi_i^\text{max})}{l_i^0}\Bigr|}_{\text{constant since } \delta_i^1=\beta}
            \label{eq:delta_dot_max}
        \end{align}
        Therefore, with well-chosen \textit{reaction distance} $v_i^0\cdot\Delta t$ and \textit{articulation threshold} $\beta$, articulation will not exceed its allowed maximum.
        \item Finally, if \emph{$|\delta_i^1|<\beta$} then $|\Delta\delta_i^1| < \max{|\Delta\delta_i^1|}$, see \labelcref{eq:delta_dot_max}.
    \end{itemize}\qedhere
\end{proof}


We choose $\beta=75\si{\degree}$ and $\max_\delta(w^\text{jack})=2$, which leaves us with $b=0.5$ and $c=2$ in \cref{eq:jackknife-weigt-func}. From \labelcref{eq:delta_dot_max}, we can then compute $v_i^\text{max}\left(\Delta t, l_i^0, l_i^1\right)$ as \labelcref{eq:vmax}.
\begin{align}
    w^\text{jack} &= 1 + \tanh(b - c \cos(\delta)) \label{eq:jackknife-weigt-func}
\end{align}
\begin{align}
    v_i^\text{max} &= \frac{\delta_i^\text{max}-\beta}{\Delta t \cdot \Bigl|\frac{\sin(\beta)}{l_i^1}+\frac{\tan(\phi_i^\text{max})}{l_i^0}\Bigr|} \label{eq:vmax}
\end{align}

To validate the applicability of the proven strategy, it is critical to assess the correctness of the assumptions.
While \cref{ass:straight-keeps-straight} is straightforward, \cref{ass:straight-reduces-jackknifing} needs to be tested experimentally; see \cref{sec:exp-results}.

\subsection{GOAL ATTRACTION}
\label{sec:goal-attraction}
Given any current pose $\{x_i^1,y_i^1,\Theta_i^0\}$ and goal pose $\mathcal{P}_i^G$, one can compute a connecting Dubins path \cite{dubinsCurvesMinimalLength1957} consisting of a concatenation of circles and up to one straight. For the radius of the circles, we choose the radius of the minimal stable circle HAV $i$ can drive on. 
\begin{theorem}
    The minimal stable circle that HAV $i$ can drive on for infinite time has radius $R_i^\text{min}$; computed by \labelcref{eq:rmin}.
    \begin{align}
        R_i^\text{min} = \sqrt{\sum_{j=0}^{N_i}(l_i^j)^2} \label{eq:rmin}
    \end{align}
\end{theorem}
\begin{proof}
    In a stable circle, all axles of the HAV rotate around a common instantaneous center of rotation (ICR). This requires that all wheels move along the tangent of their respective circles. That is, their extended axle points towards the ICR. From \cref{fig:Rmin} we get the relations ${R_i^\text{min}}^2 = {l_i^0}^2 + {R_i^1}^2$, ${R_i^1}^2={l_i^1}^2+{R_i^2}^2$, etc. until only the radius $R_i^{N+1}$ of the last axle's circle remains. Then we see that ${R_i^\text{min}}^2 = \sum_{j=0}^{N_i}(l_i^j)^2 + {R_i^{N+1}}^2$. It is trivial to see that $R_i^\text{min}$ becomes minimal when $R_i^{N+1}=0$, i.e., the last axle is in the ICR, yielding \cref{eq:rmin}.
\begin{figure}[ht]
    \centering
    \includegraphics{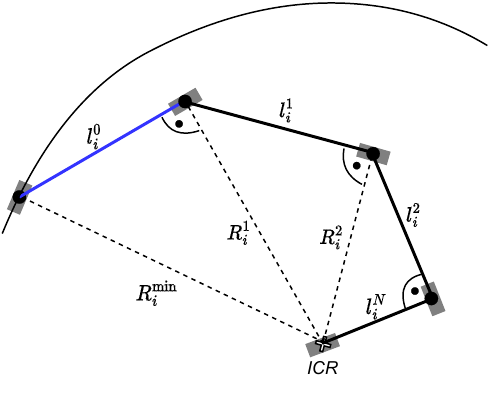}
    \caption{Minimum stable circle of an example HAV. (blue: truck, black lines: trailers, arc: sub-arc of minimal circle).}
    \label{fig:Rmin}
\end{figure}
\end{proof}

Note that the stable state might be reached only after multiple rotations of the truck on the circle, depending on starting articulation. Before that, trailer articulation will be smaller. In addition, please note that the stable state is jackknifing-free.
Also note that the HAV may transiently drive smaller-radius arcs, depending on trailer configuration and steering limits. The stable circle denotes what is jackknife-safe indefinitely, and thus safe to use in dubins paths.

To improve heading accuracy, we define an \textit{approach point} $\mathcal{P}_i^A$ one truck length before the goal $\mathcal{P}_i^G$; see \labelcref{eq:approach-point}. Then the goal heading vector $\vec{m_i^g}$ for HAV $i$ can be computed by \cref{alg:goal_attraction}.
\begin{align}
\mathcal{P}_i^A = \mathcal{P}_i^G - l_i^0 \cdot \begin{pmatrix}
    \cos{p_{\Theta,i}^G} \\ \sin{p_{\Theta,i}^G}
\end{pmatrix} \label{eq:approach-point}
\end{align}
\begin{algorithm}
\caption{Goal Attraction}\label{alg:goal_attraction}
\begin{algorithmic}
\Require $\mathcal{P}_i^G, \mathcal{P}_i^A, x_i^1,y_i^1,\Theta_i^0,R_i^\text{min}$
\Ensure $\vec{m}_i^g$
\If {close to goal}
    \State $P \gets \mathcal{P}_i^G$
    \State $R \gets l_i^0$ \Comment{Allow sharper turns when close to goal}
\Else
    \State $P \gets \mathcal{P}_i^A$ 
    \State $R \gets R_i^\text{min}$
\EndIf
\State Compute Dubins path from current pose to $P$ with $R$
\If {dubins path starts with left curve}
    \State $\phi \gets +\phi_i^\text{max}$
\ElsIf {dubins path starts with right curve}
    \State $\phi \gets -\phi_i^\text{max}$%
\EndIf
\State $\vec{m}_i^g = 1 \cdot \begin{pmatrix}
    \cos\left(\Theta_i^0+\phi\right)\\
    \sin\left(\Theta_i^0+\phi\right)
\end{pmatrix}$
\end{algorithmic}
\end{algorithm}


\subsection{MUTUAL COLLISION AVOIDANCE} 
\label{sec:collision-avoidance}
For mutual collision avoidance, we use circular approximations as described in \cref{sec:collision}. If the distance between HAVs $i$ and $h$ falls below their collision threshold $d_{i,h}$, they replace their goal attraction vectors $\vec{m}_i^g$ and $\vec{m}_h^g$ by vectors directed away from each other. Currently, priority is given to the first detected HAV that initiates evasion; cf. \labelcref{eq:jackknife-vector-addition-tmp}. Alternative policies are not explored in this paper.
This approach can similarly be applied to environmental obstacles detected during execution by modeling them as point masses with a defined collision distance that vehicles must avoid.


\section{EXPERIMENT DESIGN}
\label{sec:exp-design}
In this paper, we focus our analysis on the bilateral interaction between two swarm members, as the combination of their large state spaces leads to highly complex scenarios. To investigate these interactions, we designed a set of randomized experiments per \cref{sec:scenario} involving one and two HAVs. We test two \textbf{Hypotheses}: 1) \cref{ass:straight-reduces-jackknifing} holds, at least for HAVs with few trailers; 2) HAVs reach their goals. To evaluate Hypothesis 1, we introduce two metrics: the \textit{Joint Position Metric}, locating the joint along the HAV length, and the \textit{Length Difference Metric}, the cumulative trailer-length difference normalized by HAV length.

To model real-world truck-trailer combinations, each HAV $i$ is composed of a single truck and $N_i$ trailers, where $N_i\in\{1,\dots,10\}$ is sampled from a Rayleigh distribution with $\sigma=3$ (see \cref{fig:exp-params-Ni}). The truck length $l_i^0\in[2\si{\meter},12\si{\meter})$ is sampled from a mixed Gaussian distribution with $\mu_1=4\si{\meter},\sigma_1=0.6,\mu_2=10.7\si{\meter},\sigma_2=1.2$ (see \cref{fig:exp-params-li0}). Each trailer length $l_i^j\in[2\si{\meter},12\si{\meter})$ is independently sampled from a uniform distribution.
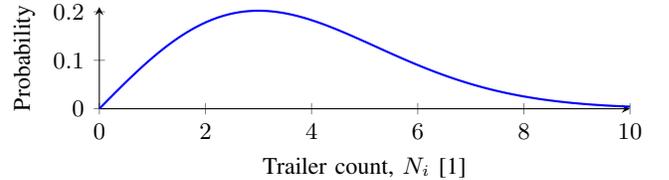
\begin{figure}[ht]
    \centering
    \begin{tikzpicture}
    \begin{axis}[
        axis lines = left,
        xlabel={Trailer count, $N_i$ [1]},
        ylabel={Probability},
        domain=0:10,
        samples=100,
        ymin=0,
        ymax=0.21,
        ytick={0,0.1,0.2},
        height=.34\linewidth,
        width=\linewidth
    ]
    \addplot [
        thick, blue
    ]
    {(x/(3^2)) * exp((-x^2)/(2*3^2))};
    \end{axis}
    \end{tikzpicture}
    \caption{Trailer Count Distribution.}
    \label{fig:exp-params-Ni}
\end{figure}
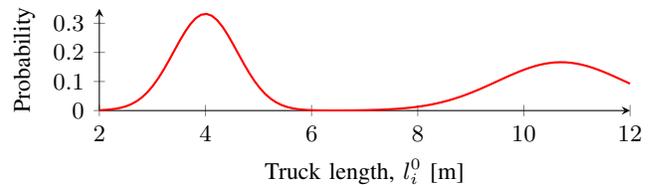
\begin{figure}[ht]
    \centering
    \begin{tikzpicture}
    \begin{axis}[
        axis lines = left,
        xlabel={Truck length, $l_i^0$ [\si{\meter}]},
        ylabel={Probability},
        domain=2:12,
        samples=100,
        ymin=0,
        ymax=0.35,
        height=.34\linewidth,
        width=\linewidth
    ]
    \addplot [
        thick, red
    ]
    {0.5 * 1/(sqrt(2*pi*0.6^2)) * exp(-(x-4)^2/(2*0.6^2)) +
     0.5 * 1/(sqrt(2*pi*1.2^2)) * exp(-(x-10.7)^2/(2*1.2^2))};
    \end{axis}
    \end{tikzpicture}
    \caption{Truck Length Distribution.}
    \label{fig:exp-params-li0}
\end{figure}
\begin{figure*}[htb]
    \centering
    \includegraphics{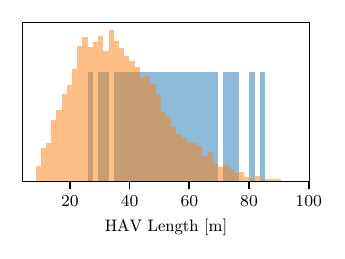}
    \hfill
    \includegraphics{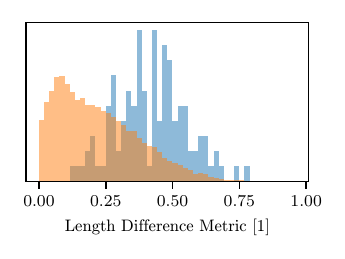}
    \hfill
    \includegraphics{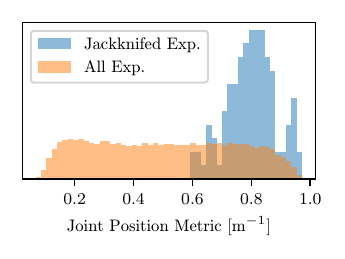}\vspace{-0.3cm}
    \caption{Probability Distribution of Experimental Results for Two HAV Experiments. Y-axis scaled differently between subplots for maximum clarity.}
    \label{fig:jackknife_2-truck_combined}
\end{figure*}

We then sequentially assign a start pose to each HAV. The pose is randomly sampled from a uniform distribution in 2D space plus heading such that $0\leq x\leq A$, $0\leq y\leq A$ and $0\leq\Theta^0\leq2\pi$, where $A^2$ is the \textit{area size} and $\forall j: \Theta^j=\Theta^0$. If the sampled pose is in potential collision with any previously sampled HAV (see \labelcref{eq:potential-collision}), a new random pose is chosen until a valid pose is found. $A$ is chosen such that the area is equal to quadruple the sum of the area footprint of all HAVs: $A^2 = 4\cdot\sum_i(\pi d_i^2)$.
The first and second goal poses of each HAV are then set following the same schema. Note that, e.g., the first goal poses are restrained with respect to each other, and there are no constraints to other phases.

A goal is determined as reached (hit) when the corresponding HAV's truck's rear axle is less than $d_e^G = \SI{0.5}{\meter}$ away: $||p_{x,i}^G-x_i^1,p_{y,i}^G-y_i^1||_2 < d_e^G$ and their heading differs no more than $d_h^G = \SI{0.1}{\radian}$: $|((p_{\Theta,i}^G-\Theta^0_i+\pi)\mod2\pi)-\pi| < d_h^G$. The HAV's maximum velocity is then imitatively reduced to zero. Once every HAV has reached its first goal, all HAVs' maximum velocities are reset, and they are assigned their second goal. For our chosen parameters, we can observe that $v_i^\text{max}\geq1\si{\meter\per\second}$ for $\Delta t=\SI{0.2}{\second}$; see \labelcref{eq:vmax}. Therefore, we set $v_i^\text{max}=1\si{\meter\per\second}$ for all trucks in our simulations. We then simulate 4,500 single-HAV and 4,500 two-HAV experiments. Each experiment runs for a maximum of 20,000 time steps, terminating early only when the second goals are reached by all HAVs in the experiment.

The code used to execute the experiments in simulation and the resulting data are provided through Zenodo\footnote{Code DOI: \href{https://doi.org/10.5281/zenodo.17342236}{10.5281/zenodo.17342236}}.


\section{SIMULATED EXPERIMENT RESULTS}
\label{sec:exp-results}
This section presents the results of the experiments outlined in \cref{sec:exp-design}. We focus on jackknifing avoidance in \cref{sec:results-jackknifing}, followed by an analysis of goal-reaching performance, reasons for missed goals, and thoughts on mutual collisions in \cref{sec:results-reachGoals,sec:results-missAnal,sec:results-collision}.

\subsection{JACKKNIFING}
\label{sec:results-jackknifing}
Out of 4,500 single-HAV experiments, only 7 (\SI{0.16}{\percent}) featured at least one jackknifed pose (i.e., violating \labelcref{eq:cos_delta_lim}). In the 4,500 two-HAV experiments (9,000 HAVs), 95 HAVs (\SI{1.1}{\percent}) experienced jackknifing.

\cref{fig:jackknife_2-truck_combined} shows three histograms for the two-HAV experiments, comparing all runs to jackknifing cases. The left histogram reveals that jackknifing is more frequent in longer HAVs, suggesting a positive correlation with vehicle length. The same trend appears for trailer count, where we didn't see any jackknifing for HAVs with less then four trailers. The second histogram, using the Length Difference Metric, indicates higher jackknifing likelihood with larger cumulative trailer-length differences, whereas the length difference at the affected joint itself does not correlate. The third histogram, showing the Joint Position Metric, reveals that jackknifing predominantly occurs at the rear joints.

These results partially support Hypothesis 1: While \cref{ass:straight-reduces-jackknifing} might not be universally valid, it holds for shorter HAVs, those with fewer trailers, and those with smaller trailer length differences. We also learn that jackknifing occurs more often during HAV interactions, suggesting a link to our mutual collision avoidance strategy.
%
%
\begin{figure*}[ht]
    \centering
    \includegraphics{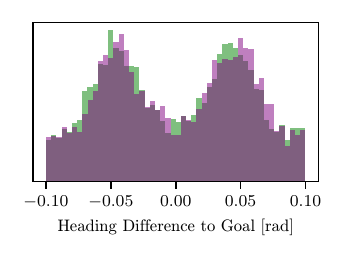}
    \hfill
    \includegraphics{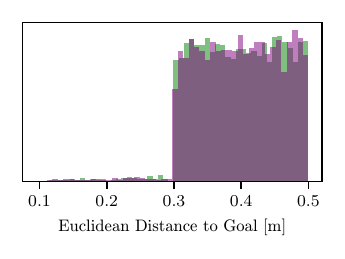}
    \hfill
    \includegraphics{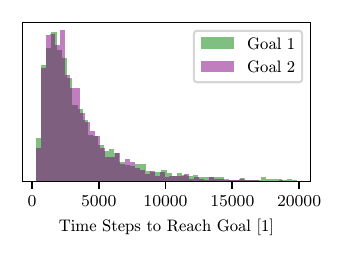}\vspace{-0.3cm}
    \caption{Probability Distribution on Goal States for Two HAV Experiments. Y-Axis scaled differently between subplots for maximum clarity.}
    \label{fig:goal-hit_2-truck_combined}
\end{figure*}
\begin{figure*}[ht]
    \centering
    \includegraphics{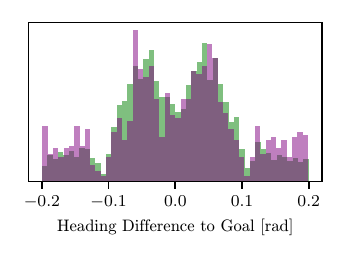}
    \hfill
    \includegraphics{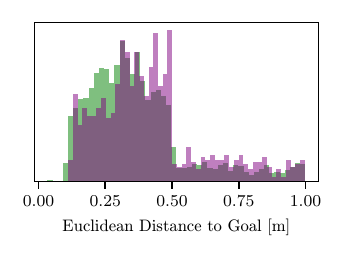}
    \hfill
    \includegraphics{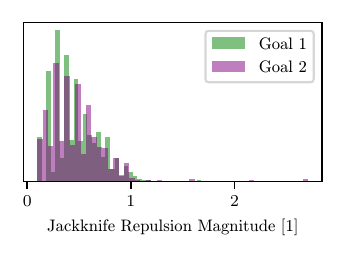}\vspace{-0.3cm}
    \caption{Probability Distribution on Close Misses for Two HAV Experiments.  Y-Axis scaled differently between subplots for maximum clarity.}
    \label{fig:goal-miss_2-truck_combined}
\end{figure*}
\subsection{GOAL REACHING PERFORMANCE}
\label{sec:results-reachGoals}
We next evaluate the HAVs' ability to reach their assigned goals. Each HAV had two sequential goals per experiment. In the 4500 single-HAV experiments, 3904 reached the first goal (\SI{86.7}{\percent}), and 3754 the second (\SI{83.4}{\percent}). In the 4500 two-HAV experiments, 7142 trucks reached the first goal (\SI{79.4}{\percent}), and 5856 the second (\SI{65.1}{\percent}).

\cref{fig:goal-hit_2-truck_combined} presents three histograms for the two-HAV experiments: heading difference at goal achievement, Euclidean distance at goal achievement (most successes \SI{0.3}{\meter}–\SI{0.5}{\meter}, threshold \SI{0.5}{\meter}), and time steps to reach each goal.

Further analysis reveals that the probability of goal achievement is independent of initial Euclidean distance, heading difference, and inter-goal distance. HAVs approach goals from a wide range of headings and finish within the allowed distance thresholds. Euclidean distance would reduce in the next simulation step, suggesting heading, not distance, limits goal completion.

Comparing the first and second goals, we find similar distributions in all metrics. 
Finally, we observe that single- and two-HAV experiments follow comparable trends, though two-HAV success rates are lower, likely due to increased complexity and collision avoidance demands.

These findings support Hypothesis 2, demonstrating that HAVs can generally reach their assigned goals, although the presence of other HAVs can reduce the overall success rate.

\subsection{GOAL MISS ANALYSIS}
\label{sec:results-missAnal}
To better understand why HAVs sometimes fail to reach their goals, we define two types of \textit{close misses}, which exclude hits. A heading close miss occurs when the Euclidean distance threshold is met and the heading difference is in $[d_h^G,2d_h^G)$. A Euclidean close miss occurs when the heading difference threshold is met and the Euclidean distance is in $[d_e^G,2d_e^G)$. Any state where either of the above criteria is true for either of the goals is recorded as a close miss. 

In these cases, of the two-HAV experiments, HAVs often experienced active jackknife repulsion; see \cref{fig:goal-miss_2-truck_combined} rightmost. However, due to static repulsion increasing with trailer count, we cannot conclusively attribute missed goals to jackknife repulsion, though some correlation is likely.

Analysis of the two leftmost histograms in \cref{fig:goal-miss_2-truck_combined} shows that close misses usually fall within the distance threshold $d_e^G$, while heading differences are more dispersed. Further investigation shows that heading close misses constitute the majority with \SI{67}{\percent} and \SI{68}{\percent} for the first and second goals.

Finally, from the previous subsection we find that two-HAV experiments have significantly more missed goals than single-HAV runs, implying mutual collision avoidance is a key factor. Yet it was inactive during close misses, suggesting earlier activation would have driven the system into a state preventing goal achievement.

%
\subsection{MUTUAL COLLISIONS}
\label{sec:results-collision}
Among the 9000 HAVs in the two-HAV experiments, a potential collision was detected 8.516 times (\SI{94.6}{\percent}), while an actual collision occurred in 26 cases (\SI{0.3}{\percent}). This means that \SI{0.3}{\percent} of potential collisions resulted in an actual collision.
One possible explanation for these actual collisions is the current prioritization of jackknife avoidance over mutual collision avoidance; cf. \labelcref{eq:jackknife-vector-addition-tmp}. To further reduce the occurrence of collisions, additional analysis is needed, including comparisons with more advanced behavior selection mechanisms such as context steering~\cite{frayContextSteeringBehaviorDriven2019}.


\FloatBarrier
\section{CONCLUSION AND FUTURE WORK}
In this work, we proposed a core mechanism for swarm behavior to reliably automate a fleet of HAVs. We present mechanisms to avoid the main problems of jackknifing and mutual collisions. Jackknifing occurred in \SI{0.2}{\percent} and \SI{1.1}{\percent} of single- and two-HAV experiments, respectively, while mutual collision was present in \SI{0.3}{\percent} of cases. Therefore, \textit{AVOID-JACK} was able to avoid both problems most of the time. Consequently, it provides a baseline for future work and establishes the foundation of this new research field.

In future work, we aim to extend our approach in several key directions. First, we plan to address more challenging environments that feature restricted spaces and known static obstacles. For collision avoidance, we intend to develop more space-efficient footprint approximations and explore methods to combine multiple constraints, such as context steering~\cite{frayContextSteeringBehaviorDriven2019,dockhornEvolutionaryAlgorithmParameter2023} or sparse roadmap spanners~\cite{honigTrajectoryPlanningQuadrotor2018}. Improving mutual collision avoidance, particularly in scenarios involving multiple simultaneous potential collisions, remains a priority, as this will allow to scale the approach to larger fleets.
Additionally, incorporating hardware experiments with real robots will be essential to validating our methods under real-world conditions, including off-axle hitching. Finally, we plan to account for vehicle-dynamics constraints, such as the inertia of trucks and limited acceleration regarding steering and longitudinal movement, to further improve the realism and applicability of our approach.
\pagebreak
\addtolength{\textheight}{-12.3cm}   


\bibliography{myBetterBibTex}

\end{document}